%% file: root.tex
\documentclass[letterpaper, 10 pt, conference]{IEEEconf}  

\IEEEoverridecommandlockouts                              
\usepackage{graphics} 
\usepackage{subcaption}
\usepackage{commath}
\usepackage{float}
\usepackage{array}
\usepackage{soul}
\usepackage{multicol}
\usepackage{epsfig} 
\usepackage{times} 
\usepackage{transparent}
\usepackage{amsmath} 
\usepackage{amssymb}  
\usepackage{amsfonts}   
\usepackage{makecell}  
\usepackage{cite}
\usepackage{bm}         
\usepackage{comment}
\usepackage[normalem]{ulem}
\usepackage{xcolor}
\usepackage{siunitx}
\usepackage{lipsum}
\newcommand\wrong[1]{\textcolor{red}{#1}}  
\newcommand\cor[1]{\textcolor{blue}{#1}}  
\graphicspath{{./}{images/}}
\usepackage{xspace}
\providecommand{\FullStop}{\text{~\@.\xspace}}
\providecommand{\Comma}{\text{~,\xspace}}       
\usepackage{todonotes}  %
\usepackage{import}
\title{\LARGE \bf
Partial Feedback Linearization Control of a  Cable-Suspended Multirotor Platform for Stabilization of an Attached Load
}

\author{Hemjyoti Das$^{1}$ and Christian Ott$^{1, 2}$
\thanks{$^{1}$Automation and Control Institute (ACIN), TU Wien, Gusshausstraße 27-29, 1040, Vienna, Austria
{\tt\small \{hemjyoti.das, christian.ott\}@tuwien.ac.at} }%
\thanks{$^{2}$Institute of Robotics and Mechatronics, German Aerospace Center (DLR), Muenchener Strasse 20, 82234, Wessling, Germany}
\thanks{This work was supported by the Austrian Research Promotion Agency (FFG) within the Program Expedition Zukunft through the HängMan Project with the FFG Project number FO999913217.}}

\begin{document}

\maketitle
\thispagestyle{empty}
\pagestyle{empty}

\begin{abstract}
\import{sections/}{0-abstract}
\end{abstract}

\section{Introduction}
\import{sections/}{1-introduction}

\section{System Modelling}
\import{sections/}{3-system_dynamics}

 \section{Partial Feedback Linearization}
\import{sections/}{4-pfl}





\section{Simulation Results}
\import{sections/}{8-simulations}

\section{Experimental Results}
\import{sections/}{9-experiments}

\section{Conclusion and Future Work}
\import{sections/}{10-conclusion}

\bibliographystyle{IEEEtran}
\bibliography{bibliography}{}

\end{document}

%% file: sections/0-abstract.tex
 In this work, we present a novel control approach based on partial feedback linearization (PFL) for the stabilization of a suspended aerial platform with
an attached load. Such systems are envisioned for various
applications in construction sites involving cranes, such as the
holding and transportation of heavy objects. Our proposed
control approach considers the underactuation of the whole system while utilizing its coupled dynamics for stabilization. We demonstrate using numerical stability analysis that these coupled terms are crucial for the stabilization of the complete system. We also carried out robustness analysis of the proposed approach in the presence of external wind disturbances, sensor noise, and uncertainties in system dynamics. As our envisioned target application involves
cranes in outdoor construction sites, our control approaches
rely on only onboard sensors, thus making it suitable for such
applications. We carried out extensive simulation studies and
experimental tests to validate our proposed control approach.

%% file: sections/1-introduction.tex
In recent years, aerial robotic manipulation has grown significantly and has been used for various applications such as inspection of pipelines and other remote sites, transportation of payloads, and aerial mapping \cite{ollero2018aeroarms,  ollero2021past, staub2018towards, gonzalez2024controlled}. The total flight time and the payload capacity are often limited due to the excessive energy consumed for the stabilization of such platforms \cite{paredes2017study,villa2020survey}. Additionally, it can raise safety concerns and turbulence in the surroundings due to the large size of the propellers necessary to support heavy manipulator \cite{kondak2014aerial}. 
To overcome these issues, the cable-suspended aerial manipulator (SAM) is proposed in \cite{sarkisov2019development}, where an overhead crane system compensates for the weight of both the platform and the manipulator. \par 
This platform is utilized for various applications, such as in  \cite{gabellieri2020compliance} to demonstrate compliant interaction with the environment, while in \cite{coelho2021hierarchical}, a hierarchical control framework is utilized to enhance its field of view (FoV) from its onboard camera for efficient teleoperation. The pendulum motion generated due to its suspension is one of its main challenges, which has been addressed in \cite{sarkisov2020optimal} using an optimal feedback control law. In \cite{yiugit2020preliminary, yiugit2021novel}, a similar suspended omnidirectional platform is proposed, where a nonlinear model predictive controller (NMPC) is utilized to perform accurate trajectory tracking, while utilizing the elasticity of its suspension. Similarly, various teleoperation tasks have been demonstrated in  \cite{kong2022robotic, perozo2024teleoperation} with such suspended aerial manipulator platforms. 
\begin{figure}[t]
    \centering
\includegraphics[width=1\linewidth]{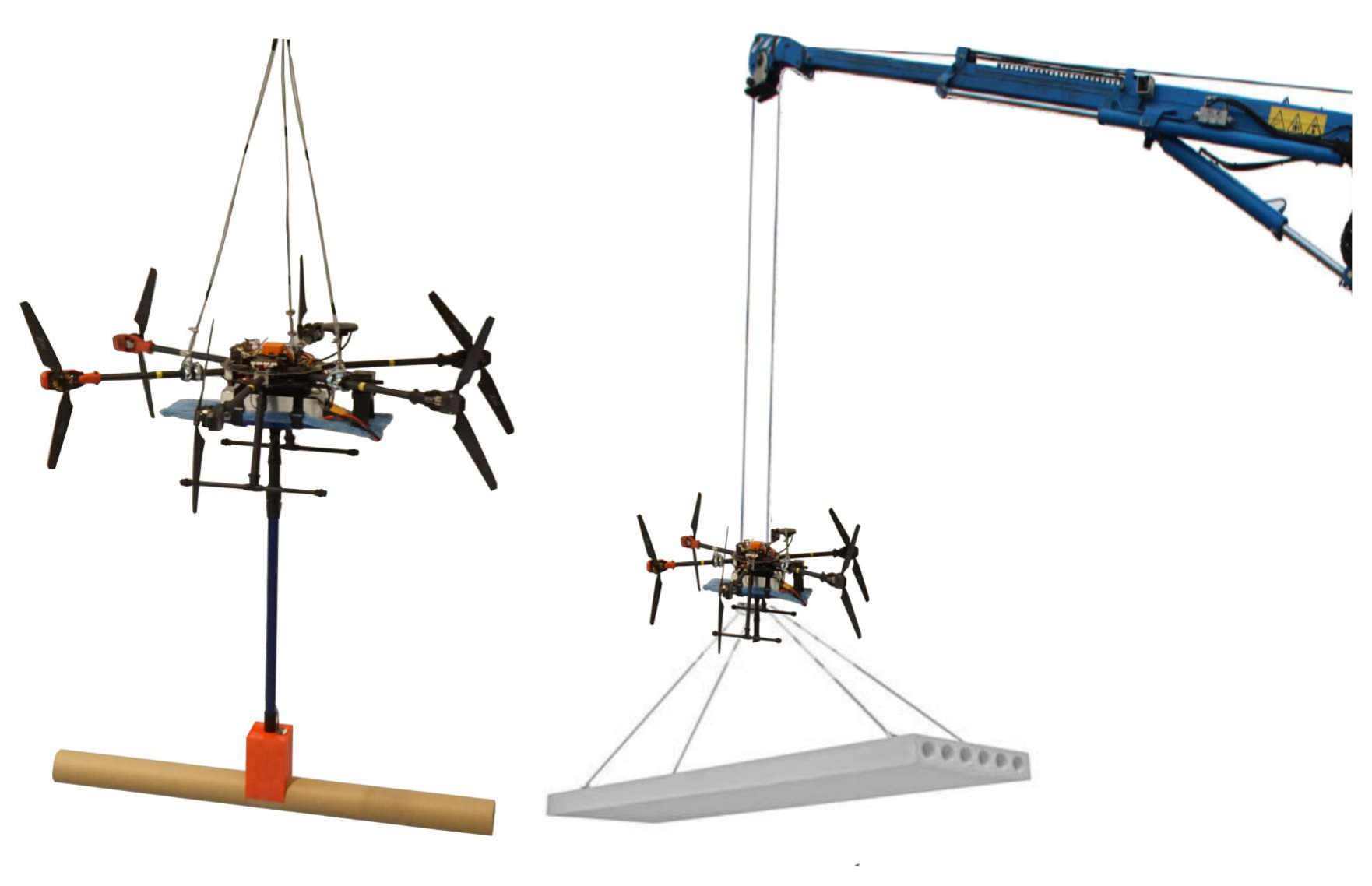}  
    \caption{{Cable-suspended multirotor platform with an attached load used for experimental validation} (left) and {the envisioned target application of load transportation using cranes in construction sites (right).}}
    \label{fig:intro}
\end{figure}
\par 
\par 
This work extends the application of such a suspended multirotor platform for the transportation of heavy loads, similar to our previous work \cite{das2024}. Handling slung loads using such a suspended multirotor platform differs from that of the conventional free-flying multirotor platforms, which has been widely studied in the literature. 
For instance, a differential-flatness based controller is designed in \cite{sreenath2013geometric} for a quadrotor-load system, which is not easy to design for underactuated systems with passive joints \cite{he2016characteristics}. Furthermore, such a controller relies on the accuracy of the system model, and thus necessitates certain adaptation to accommodate any involved uncertainties \cite{poultney2018robust}.
In \cite{chang2023new}, an adaptive controller ensures stabilization of such a system but is computationally complex and lacks hardware validation, whereas in \cite{klausen2017nonlinear}, a backstepping controller is utilized to mitigate disturbances robustly in such systems. \par 
 The payload capacity and the total flight time in such free-flying aerial platforms are limited by their energy constraints, which can restrict their effectiveness in many applications. A suspended aerial platform, however,  can significantly compensate for these limitations. Consequently, this demands different stabilization methods for the entire system due to the suspension of the multirotor platform from a crane. Although they need specialized control, these platforms offer significant advantages in load stabilization. In particular, they can complement existing load stabilization techniques used in overhead crane systems, such as the energy shaping control in \cite{wu2016nonlinear}, which ensures minimal payload swing. The approach in \cite{wu2016nonlinear} relies on the precision of the system model, and thus inaccurate modeling of the crane dynamics can lead to instability of the system. In  \cite{ouyang2018sliding}, a sliding mode controller is utilized for load stabilization from overhead cranes, while in 
\cite{wu2020disturbance}, a nonlinear disturbance observer is utilized to compensate for uncertain disturbances in real-time. 
\par 
The addition of a multirotor platform in these crane-based systems will help compensate for model inaccuracies of the crane in the designed controller. It will also share the burden of load stabilization with the crane, reducing the wear and tear on its mechanical structure. Additionally, due to the faster response time of the multirotors, the total active sway suppression can be greatly improved in the presence of external disturbances. In our previous work \cite{das2024}, we have tackled a similar load stabilization problem from a suspended multirotor platform using a composite control approach, based on the assumption of time-scale separation between different involved dynamics of the system.  However, this assumption is not made in the present work. Furthermore, unlike in our previous work, we now consider the platform to be directly attached to the crane using cables, without using an additional hook to mount the platform, thus eliminating its associated oscillatory dynamics as observed in \cite{das2024} and \cite{sarkisov2023hierarchical}. \par 
Building on these points, we summarize the main contributions of this paper as follows:
\begin{itemize}
    \item Design of a novel stabilization controller for a cable-suspended multirotor platform with an attached load. Our approach is based on partial feedback linearization (PFL) to stabilize the complete system. 
    \item We demonstrate the robustness of our control approach by introducing sensor noise, external wind disturbances, and uncertainties in the system model. 
    \item We demonstrate the numerical stability of our approach in comparison with the standard PFL approach. 
    \item We validate our approach both by using an extensive simulation study and experimental evaluation.
\end{itemize} \par 
 Similar to our previous works in \cite{das2023, das2024}, we rely on only onboard sensors for state-estimation and thus making it suitable for use in different construction sites for transportation of an attached load. Additionally, our control approach is also capable of stabilizing the platform-load system at non-equilibrium points.\par 
The rest of the paper is organized as follows. In Section \ref{sec:dynamics}, we describe the dynamics of the complete system, following which in Section \ref{sec:PFL}, we present our approach of PFL control, while demonstrating its numerical stability. In Section \ref{sec:sim}, we present our simulation results for the nominal case while testing the robustness of our proposed approach. Then in Section \ref{sec:expm}, we present the experimental results, and finally, the conclusion and recommendations for future work are drawn in Section \ref{sec:con}. 

%% file: sections/3-system_dynamics.tex
\begin{figure}[t]
    \centering
    \def\svgwidth{0.8\columnwidth}
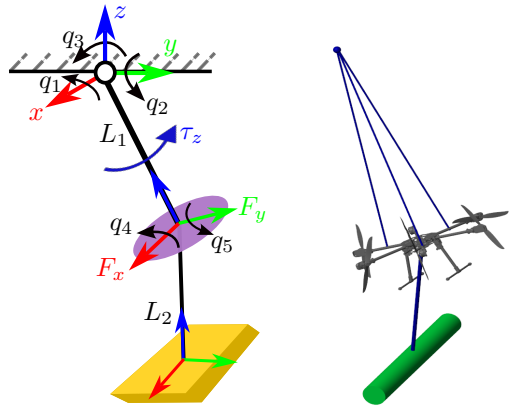
    \caption{Schematic diagram of the cable-suspended multirotor platform with the attached load (left) and stabilization of the simulated system about non-equilibrium points $q_1 =$ \SI{-0.1}{\radian}, $q_2 =$ \SI{-0.5}{\radian}, $q_3 =$ \SI{0}{\radian}, $q_4 =$ \SI{0.1}{\radian} and $q_5=$ \SI{0.5}{\radian} (right). }
    \label{fig:platform_schematics}
\vspace{-10pt}
\end{figure}

\label{sec:dynamics}
In this section, we present the mathematical model of the cable-suspended multirotor platform with the attached load. The platform is modeled as a spherical pendulum with 3 DOFs because its suspension allows unrestricted rotation and swinging around the attachment point. The attached load, on the other hand, is modeled as a spherical pendulum with 2 DOFs, as it primarily swings due to gravity but lacks independent rotational freedom about the cable, due to mechanical constraints in the suspension mechanism. Together, the platform and the load constitute a double spherical pendulum with 5 DOFs, where the platform is suspended from a fixed point (see Fig. \ref{fig:platform_schematics}) and the load is attached to the platform, both using rigid cables. Moreover, the cables used for the suspension are assumed to be taut, similar to \cite{sarkisov2019development, gabellieri2020compliance}. \par 
The first spherical joint involves the rotation of the suspension cable with length $L_1$ about its $x$, $y$, and $z$ axes, denoted as $q_1$, $q_2$, and $q_3$, respectively. The second spherical joint involves the rotation of the suspension cable with length $L_2$ about its $x$ and $y$ axes, denoted by the angles $q_4$ and $q_5$, respectively. The system dynamics of the complete system can be summarized as 
\begin{equation}
\label{eq:dyn}
    \mathbf{M\left(q\right)}\mathbf{\ddot{q}} +  \mathbf{C\left(q, \dot{q}\right)}\mathbf{\dot{q}} +  \mathbf{g} =  \bm{\tau} \Comma
\end{equation} 
where $\mathbf{q} = \begin{bmatrix} q_1 & q_2 & q_3 & q_4 & q_5 \end{bmatrix}^T$ is the joint configuration vector, $\mathbf{M(q)}\in\mathbb{R}^{5 \times 5}$ is the inertia matrix, $\mathbf{C(q, \dot{q})}\in\mathbb{R}^{5 \times 5}$ is the centrifugal/Coriolis matrix, $\mathbf{g(q)}\in\mathbb{R}^{5}$ is the gravity vector and $\mathbf{\bm{\tau}}\in\mathbb{R}^{5}$ is the joint torque vector. The torques corresponding to the attached load are zero, as the joints $q_4$ and $q_5$ serve as the passive joints of the system. The joint torque $\bm{\tau}$ can be expressed as 
\begin{equation}
    \bm{\tau} = \mathbf{J}_u^T\mathbf{u}
    \Comma
\end{equation} 
where $\mathbf{J}_u \in \mathbb{R}^{3 \times 5}$ is the Jacobian relating the applied wrench to all the joints of the system, and the wrench vector $\mathbf{u} \in \mathbb{R}^{3}$ is defined as 
\begin{equation}
\begin{aligned}
\mathbf{u} &= \begin{bmatrix}
        F_x & F_y & \tau_z \end{bmatrix}^{T}  \FullStop
\end{aligned}
\end{equation} \par 
 The suspended platform used in this work, as in our previous work \cite{das2023}, can exert the planar wrench $\mathbf{u} $ due to the adjustment of the installation angles of its unidirectional thrusters to  $\pm 90 $ degrees about the radial axis of the platform. This planar wrench
comprises the translational force along its $x$ and $y$ axis, denoted as $F_x$ and $F_y$, respectively, and the torque $\tau_z$ exerted about its $z$ axis. 

%% file: images/platform_schematics_4.pdf_tex
\begingroup%
  \makeatletter%
  \providecommand\color[2][]{%
    \errmessage{(Inkscape) Color is used for the text in Inkscape, but the package 'color.sty' is not loaded}%
    \renewcommand\color[2][]{}%
  }%
  \providecommand\transparent[1]{%
    \errmessage{(Inkscape) Transparency is used (non-zero) for the text in Inkscape, but the package 'transparent.sty' is not loaded}%
    \renewcommand\transparent[1]{}%
  }%
  \providecommand\rotatebox[2]{#2}%
  \newcommand*\fsize{\dimexpr\f@size pt\relax}%
  \newcommand*\lineheight[1]{\fontsize{\fsize}{#1\fsize}\selectfont}%
  \ifx\svgwidth\undefined%
    \setlength{\unitlength}{259.39918818bp}%
    \ifx\svgscale\undefined%
      \relax%
    \else%
      \setlength{\unitlength}{\unitlength * \real{\svgscale}}%
    \fi%
  \else%
    \setlength{\unitlength}{\svgwidth}%
  \fi%
  \global\let\svgwidth\undefined%
  \global\let\svgscale\undefined%
  \makeatother%
  \begin{picture}(1,0.84074021)%
    \lineheight{1}%
    \setlength\tabcolsep{0pt}%
    \put(0,0){\includegraphics[width=\unitlength,page=1]{platform_schematics_4.pdf}}%
    \put(0.22414312,0.79208156){\color[rgb]{0,0,0}\makebox(0,0)[lt]{\lineheight{1.25}\smash{\begin{tabular}[t]{l}\textcolor{blue}{$z$}\end{tabular}}}}%
    \put(0.06101384,0.59556887){\color[rgb]{0,0,0}\makebox(0,0)[lt]{\lineheight{1.25}\smash{\begin{tabular}[t]{l}\textcolor{red}{$x$}\end{tabular}}}}%
    \put(0.32000281,0.7285567){\color[rgb]{0,0,0}\makebox(0,0)[lt]{\lineheight{1.25}\smash{\begin{tabular}[t]{l}\textcolor{green}{$y$}\end{tabular}}}}%
    \put(0.28171504,0.21142225){\color[rgb]{0,0,0}\makebox(0,0)[lt]{\lineheight{1.25}\smash{\begin{tabular}[t]{l}$L_2$\end{tabular}}}}%
    \put(0.19558206,0.55289824){\color[rgb]{0,0,0}\makebox(0,0)[lt]{\lineheight{1.25}\smash{\begin{tabular}[t]{l}$L_1$\end{tabular}}}}%
    \put(0,0){\includegraphics[width=\unitlength,page=2]{platform_schematics_4.pdf}}%
    \put(0.21697269,0.38436992){\color[rgb]{0,0,0}\makebox(0,0)[lt]{\lineheight{1.25}\smash{\begin{tabular}[t]{l}$q_4$\end{tabular}}}}%
    \put(0,0){\includegraphics[width=\unitlength,page=3]{platform_schematics_4.pdf}}%
    \put(0.41119245,0.34880904){\color[rgb]{0,0,0}\makebox(0,0)[lt]{\lineheight{1.25}\smash{\begin{tabular}[t]{l}$q_5$\end{tabular}}}}%
    \put(0,0){\includegraphics[width=\unitlength,page=4]{platform_schematics_4.pdf}}%
    \put(0.28683146,0.61460256){\color[rgb]{0,0,0}\makebox(0,0)[lt]{\lineheight{1.25}\smash{\begin{tabular}[t]{l}$q_2$\end{tabular}}}}%
    \put(0.12500913,0.7432731){\color[rgb]{0,0,0}\makebox(0,0)[lt]{\lineheight{1.25}\smash{\begin{tabular}[t]{l}$q_3$\end{tabular}}}}%
    \put(0.08124549,0.65781527){\color[rgb]{0,0,0}\makebox(0,0)[lt]{\lineheight{1.25}\smash{\begin{tabular}[t]{l}$q_1$\end{tabular}}}}%
    \put(0,0){\includegraphics[width=\unitlength,page=5]{platform_schematics_4.pdf}}%
    \put(0.35193085,0.55992327){\color[rgb]{0,0,0}\makebox(0,0)[lt]{\lineheight{1.25}\smash{\begin{tabular}[t]{l}$\textcolor{blue}{\tau_z}$\end{tabular}}}}%
    \put(0.19000809,0.29586767){\color[rgb]{0,0,0}\makebox(0,0)[lt]{\lineheight{1.25}\smash{\begin{tabular}[t]{l}$\textcolor{red}{F_x}$\end{tabular}}}}%
    \put(0.46330189,0.41286761){\color[rgb]{0,0,0}\makebox(0,0)[lt]{\lineheight{1.25}\smash{\begin{tabular}[t]{l}$\textcolor{green}{F_y}$\end{tabular}}}}%
  \end{picture}%
\endgroup%

%% file: sections/4-PFL.tex
\label{sec:PFL}
The technique of partial feedback linearization (PFL) has been widely used in prior works to control underactuated systems, particularly those with passive joints.  For instance, in \cite{spong1994partial}, the control of a 2-DOF swing-up acrobat robot is analyzed where the internal dynamics of the system enter a stable limit cycle, under the influence of the control action designed for the output dynamics. Similar concepts of limit cycle of the internal dynamics are discussed in \cite{grizzle2002asymptotically, grizzle2017virtual} for the locomotion of a bipedal robot. In \cite{kim2018oscillation}, a high set of gains is used to stabilize an underactuated system using PFL, under the assumption of time-scale separation between different dynamics of the system. \par
\subsection{Standard PFL}
Building upon these prior approaches, we now focus on designing a PFL strategy for the stabilization of our given system. To achieve this, we begin by classifying the joint configuration vector $\mathbf{q}$ into three sub-classes as
\begin{equation}
    \label{eq:sub-classes}
\mathbf{q}_a = \begin{bmatrix}
    q_1 & q_2
\end{bmatrix}^T, 
\mathbf{q}_b = \begin{bmatrix}
 q_3   
\end{bmatrix}, 
\mathbf{q}_c = \begin{bmatrix}
    q_4 & q_5
\end{bmatrix}^T
\FullStop
\end{equation} \par 
Note that the joints $q_1$, $q_2$, and $q_3$ are the active joints of the system, which are actuated, while the joints $q_4$ and $q_5$ are the passive joints. The output vector $\mathbf{y}$ to be stabilized by the PFL consists of these actuated joints, summarized as 
\begin{equation}
    \mathbf{y} = \begin{bmatrix}
        \mathbf{q}_a & \mathbf{q}_b
    \end{bmatrix}^T \FullStop
\end{equation} \par
As the output vector $\mathbf{y}$ consists of only the actuated joints, we can express it as a function of the total joint-configuration vector $\mathbf{q}$ as 
\begin{equation}
\label{eq:y_b}
    \mathbf{y} = \mathbf{B}^T \mathbf{q} \Comma
\end{equation}
where the underactuation matrix $\mathbf{B} \in \mathbb{R}^{5 \times 3}$ is defined as 
\begin{equation}
\mathbf{B} = \begin{bmatrix}
    \mathbf{I}_{3 \times 3} & \mathbf{0}_{2 \times 3}
\end{bmatrix}^T \FullStop
\end{equation}  \par 
The dynamics of the output vector can then be expressed using \eqref{eq:dyn} and \eqref{eq:y_b} as 
\begin{equation}
\label{eq:dyn_y}
\begin{aligned}
   \mathbf{ \ddot{y}} =& \mathbf{B}^T \mathbf{\ddot{q}} \\ 
   =&\mathbf{B}^T\left(\mathbf{M}^{-1}  \left(\mathbf{J}_u^T\mathbf{u} - \mathbf{C}\mathbf{\dot{q}} - \mathbf{g}\right) \right) \\
    =&\mathbf{B}^T\mathbf{M}^{-1} \mathbf{J}_u^T\mathbf{u} - \mathbf{B}^T\mathbf{M}^{-1}\mathbf{C}\mathbf{\dot{q}} - \mathbf{B}^T\mathbf{M}^{-1}\mathbf{g} \FullStop
\end{aligned}
\end{equation} \par 
 We can rewrite the output dynamics \eqref{eq:dyn_y} as
\begin{equation}
\label{eq:dyn_y_pd+}
\bm{\Lambda}_y
 \mathbf{ \ddot{y}} + \bm{\mu}_y + \bm{\rho}_y = \mathbf{u} 
 \Comma
\end{equation}
where the dynamic matrices $\bm{\Lambda}_y$, $\bm{\mu}_y$, and $\bm{\rho}_y$ are the transformed inertia, centrifugal/Coriolis, and gravity component, respectively, and are obtained as
\begin{equation}
\begin{aligned}
  \label{eq:dyn_y_pd+_var}
\bm{\Lambda}_y &= \left(\mathbf{B}^T\mathbf{M}^{-1}\mathbf{J}_u^T\right)^{-1} \\ 
\bm{\mu}_y &= \bm{\Lambda}_y\mathbf{B}^T\mathbf{M}^{-1} \mathbf{C}\mathbf{\dot{q}}\\ 
\bm{\rho}_y &= \bm{\Lambda}_y\mathbf{B}^T\mathbf{M}^{-1} \mathbf{g} \FullStop
\end{aligned}
\end{equation} \par 
To control the output dynamics, we command the following wrench using the actuated joints 
\begin{equation}
    \label{eq:u_pd+}
    \mathbf{u} = \bm{\mu}_y + \bm{\rho}_y + \bm{\Lambda}_y \mathbf{v}_{y} -  \mathbf{{K}}_{dy} \mathbf{\dot{\tilde{y}}} - \mathbf{{K}}_{py} {\mathbf{\tilde{y}}} \Comma
\end{equation}  
where $
 \mathbf{v}_{y}$ = $\begin{bmatrix}
     \mathbf{v}_{a} & 
     0
 \end{bmatrix}^T$ is the control signal for the system output. The control signal $\mathbf{v}_{a} \in \mathbb{R}^2$ 
 for the joint group $\mathbf{{q}}_a$ is designed to result in the stabilization of $\mathbf{q}_c$, as will be discussed below. In \eqref{eq:u_pd+}, $\mathbf{K}_{dy}$ and $\mathbf{K}_{py}$ are the chosen symmetric and positive definite controller gain matrices, while  $\mathbf{\tilde{y}}$ and $\mathbf{\dot{\tilde{y}}}$ are the respective errors of the output states and its velocities from their given reference. \par 
Next, the dynamics of $\mathbf{q}_c$ is considered as the internal dynamics of the complete system, which can be extracted from \eqref{eq:dyn} as 
\begin{equation}
\label{eq:dyn_qc}
 \mathbf{\ddot{q}}_c + \bm{\mu}_c + \bm{\rho}_c = \bm{\lambda}_c \mathbf{u}  \Comma 
\end{equation} 
where $\bm{\mu}_c \in \mathbb{R}^{2}$, $\bm{\rho}_c\in \mathbb{R}^{2}$ and $\bm{\lambda}_c\in \mathbb{R}^{2 \times 3}$ are its transformed centrifugal/Coriolis vector, gravity vector and control effectiveness matrix, respectively,  defined as 
\begin{equation}
\begin{aligned}
    \label{eq:extract}
\bm{\mu}_c &= \mathbf{A}\mathbf{M}^{-1}\mathbf{C}\mathbf{\dot{q}} \\
\bm{\rho}_c &= \mathbf{A}\mathbf{M}^{-1}\mathbf{g}\\
\bm{\lambda}_c &= \mathbf{A}\mathbf{M}^{-1} \mathbf{J}_u^T\Comma
\end{aligned}  
\end{equation}
where the matrix $\mathbf{A} = \begin{bmatrix}
    \mathbf{0}_{2 \times 3} & \mathbf{I}_{2 \times 2}
\end{bmatrix}$ is the selection matrix. We then reformulate \eqref{eq:u_pd+} as 
\begin{equation}
\label{eq:u_2}
    \mathbf{u} = \mathbf{G}\mathbf{v}_{a} + \mathbf{R} \Comma
\end{equation}
where the matrices $\mathbf{G}$ and $\mathbf{R}$ are defined as 
\begin{equation}
    \begin{aligned}
       \mathbf{G} = & \mathbf{\Lambda}_y \begin{bmatrix}
    \mathbf{I}_{2\times2} & \mathbf{0}_{1\times2}
\end{bmatrix}^T \\ 
\mathbf{R} = & \bm{\mu}_y + \bm{\rho}_y -  \mathbf{{K}}_{dy} \mathbf{\dot{\tilde{y}}} - \mathbf{{K}}_{py} {\mathbf{\tilde{y}}}  \FullStop
    \end{aligned} 
\end{equation} \par 
 
Note that the matrix $\mathbf{G}$ consists of the components of the inertia $\bm{\Lambda}_y$ corresponding to the joint group $\mathbf{q}_a$, which will be utilized for the stabilization of the suspended load. The torque $\tau_3$ applied at the joint $q_3$ has no significant impact on the load dynamics, compared to the other actuated joints and, therefore, is not used for load stabilization. The chosen matrix $\mathbf{G}$ thus ensures strong inertial coupling between the joint groups $\mathbf{q}_a$ and $\mathbf{q}_c$, necessary for the stabilization. Building on this, by utilizing  \eqref{eq:dyn_qc} and \eqref{eq:u_2}, we obtain the following closed-loop internal dynamics of $\mathbf{q}_c$
\begin{equation}
\label{eq:dyn_qc_2}
   \mathbf{\ddot{q}}_c + \bm{\mu}_c + \bm{\rho}_c = \bm{\lambda}_c \mathbf{G}\mathbf{v}_{a} +\bm{\lambda}_c \mathbf{R}  \FullStop
\end{equation} \par 

In order to obtain a stable closed-loop internal dynamics, the reference $\mathbf{v}_{a}$ can be chosen as 
\begin{equation}
\begin{aligned}
    \label{eq:qa_ref}
    \mathbf{v}_{a} = \left(\bm{\lambda}_c \mathbf{G}\right)^{-1}\left( \bm{\mu}_c + \bm{\rho}_c - \bm{\lambda}_c \mathbf{R} 
   - \mathbf{K}_{dc}\mathbf{\dot{\tilde{q}}}_c - \mathbf{K}_{pc}\mathbf{\tilde{q}}_c\right) \Comma
    \end{aligned} 
\end{equation}  
where $\mathbf{K}_{dc}$ and $\mathbf{K}_{pc}$ are the controller gains. Note that in \eqref{eq:qa_ref}, for the inversion of $\bm{\lambda}_c \mathbf{G}$ to be possible, the number of columns in matrix $\mathbf{G}$ corresponding to actuated joints must match the number of internal dynamics. As a result, utilizing \eqref{eq:qa_ref} leads to following stable dynamics of $\mathbf{q}_c$
\begin{equation}
    \mathbf{\ddot{q}}_c +\mathbf{K}_{dc}\mathbf{\dot{\tilde{q}}}_c + \mathbf{K}_{pc}\mathbf{\tilde{q}}_c = \mathbf{0} \FullStop
\end{equation} \par 
Then, by utilizing \eqref{eq:qa_ref}, we obtain the following closed-loop dynamics of the output $\mathbf{y}$
\begin{equation}
\bm{\Lambda}_y\mathbf{\ddot{{y}}} +\mathbf{K}_{dy}\mathbf{\dot{\tilde{y}}} + \mathbf{K}_{py}\mathbf{\tilde{y}} =  \mathbf{G} \mathbf{v}_{a} \FullStop
\end{equation}
\par

We observe that the control signal $\mathbf{{v}}_{a}$ does not converge to zero, even when the internal states reach zero,  due to the contribution of its various nonlinear terms. This leads to a limit cycle of the output dynamics, as will be discussed in the subsequent section. Such behavior is undesirable, as our goal is to stabilize the complete system consisting of both the suspended platform and the attached load.

\subsection{PFL with Coupling}
In order to tackle the aforementioned issue, we propose the control signal $\mathbf{{\bar{v}}}_{a}$ instead of $\mathbf{{{v}}}_{a}$ as follows
\begin{equation}
\begin{aligned}
    \label{eq:qa_ref_bar}
    \mathbf{{\bar{v}}}_{a} = \mathbf{{v}}_{a} + \left(\bm{\lambda}_c \mathbf{G}\right)^{-1} \bm{\lambda}_c\left(- \mathbf{K}_{dy}\mathbf{\dot{\tilde{y}}} - \mathbf{K}_{py}\mathbf{{\tilde{y}}}\right)
    \FullStop
    \end{aligned}
\end{equation} \par 
The selection of the control law \eqref{eq:qa_ref_bar} results in the following closed-loop system dynamics
\begin{equation}
\label{eq:y_closed_loop}
\begin{aligned}
    \mathbf{\ddot{q}}_c +\mathbf{K}_{dc}\mathbf{\dot{\tilde{q}}}_c + \mathbf{K}_{pc}\mathbf{\tilde{q}}_c &= \mathbf{N}_c\\
 \bm{\Lambda}_y\mathbf{\ddot{{y}}} +\mathbf{K}_{dy}\mathbf{\dot{\tilde{y}}} + \mathbf{K}_{py}\mathbf{\tilde{y}} &=  \mathbf{N}_y \Comma
\end{aligned}
\end{equation}
where the vectors $\mathbf{N}_c \in \mathbb{R}^2 $ and $\mathbf{N}_y \in \mathbb{R}^3$ are defined as 
\begin{equation}
\begin{aligned}
    \mathbf{N}_c &= \bm{\lambda}_c \left(- \mathbf{K}_{dy}\mathbf{\dot{\tilde{y}}} - \mathbf{K}_{py}\mathbf{{\tilde{y}}}\right) \\
    \mathbf{N}_y &=\bm{\Lambda}_y  \begin{bmatrix}
        \mathbf{\bar{v}}_{a} & 0
    \end{bmatrix}^T \FullStop
\end{aligned}  
\end{equation} \par 
\begin{figure}[t]
\begin{subfigure}{.5\textwidth}
  \centering
\includegraphics[width=0.925\linewidth]{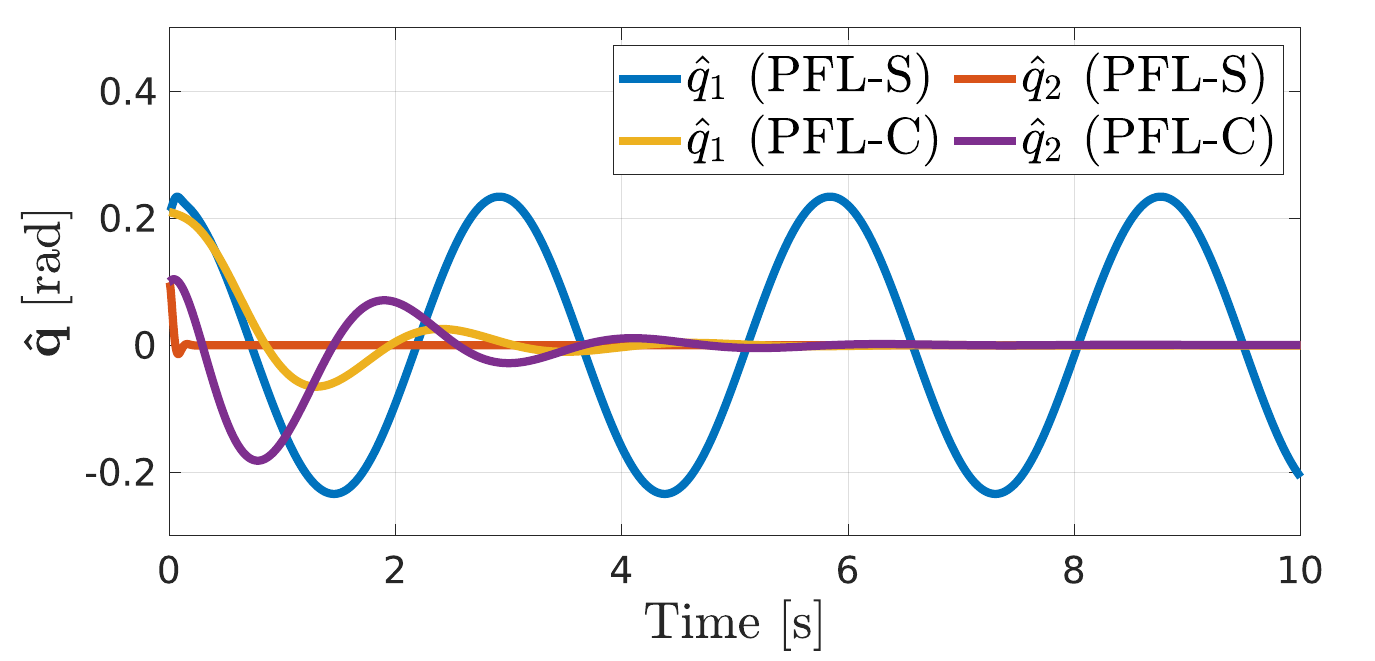}  
  \caption{}
  \label{fig:2DOF_system_beh}
\end{subfigure}
\begin{subfigure}{.5\textwidth}
  \centering
\includegraphics[width=1\linewidth]{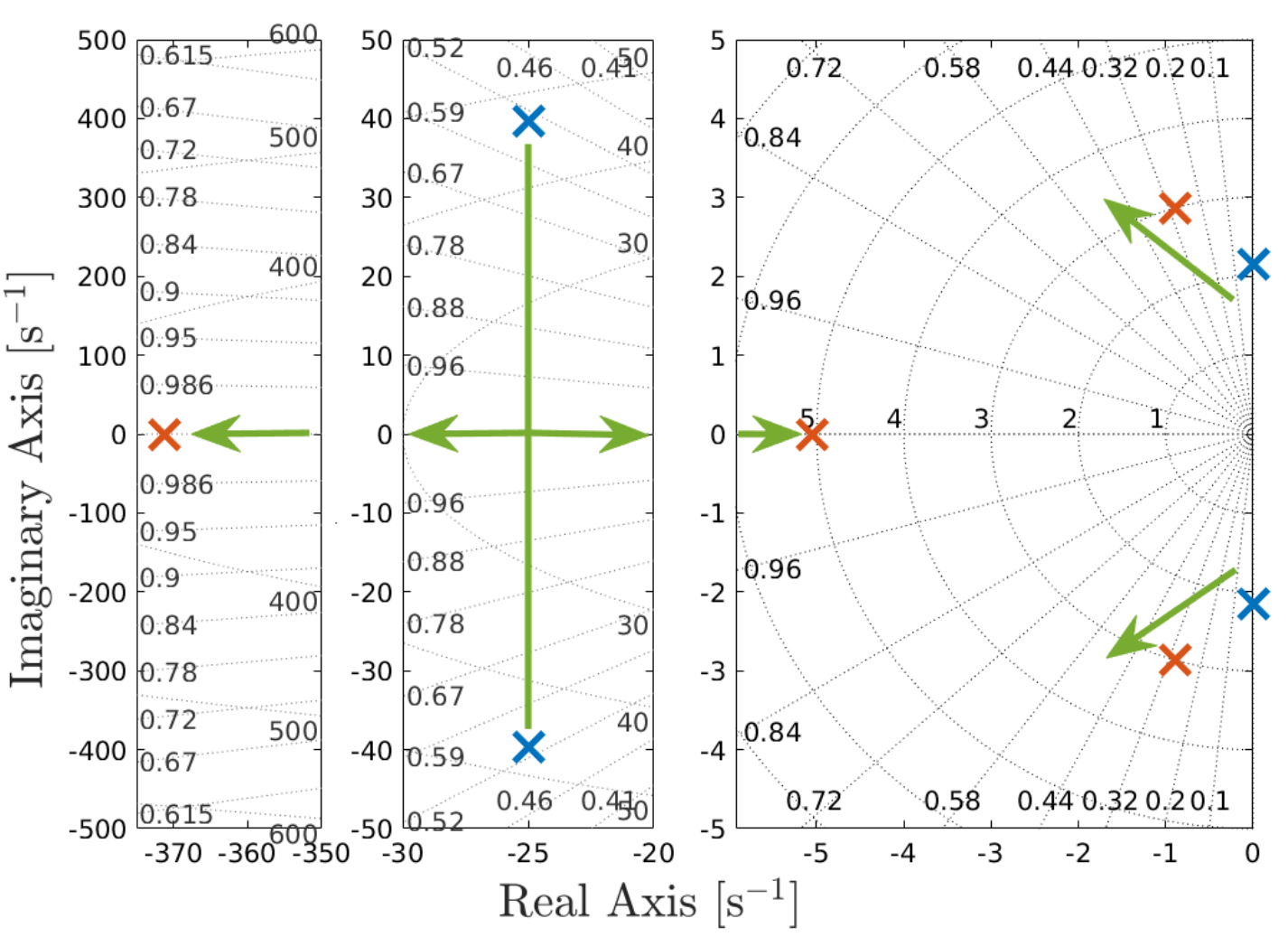} 
\vspace{-15pt}
  \caption{}
  \label{fig:pfl_com-2}
\end{subfigure}
\caption{(a) System behavior of the 2-DOF planar case using our proposed PFL approach with coupling (represented by PFL-C) and using standard PFL approach (represented by PFL-S), and b) eigenvalues of our proposed approach (shown by orange crosses) and the standard PFL approach (shown by blue crosses). The shift of the eigenvalues occurring due to the coupled terms in our proposed approach is indicated by the green arrow.}
\label{fig:sim_pfl_comp}
\vspace{-7pt}
\end{figure}
Both $\mathbf{N}_c$ and $\mathbf{N}_y$ comprise the coupling terms that are crucial for the stabilization of the complete system, which cannot be entirely eliminated from the closed loop dynamics due to the involved underactuation. In the subsequent section, we perform the stability analysis of this control approach, demonstrating the essential role these coupling terms play in ensuring system stability. The applied wrench for our proposed approach is finally summarized as 
\begin{equation}
\label{u:proposed}
    \mathbf{u} = \mathbf{G}\mathbf{\bar{v}}_{a} + \mathbf{R} 
\end{equation}
\par

\begin{figure*}[t]
\begin{subfigure}{.5\textwidth}
  \centering
  \includegraphics[width=1\linewidth]{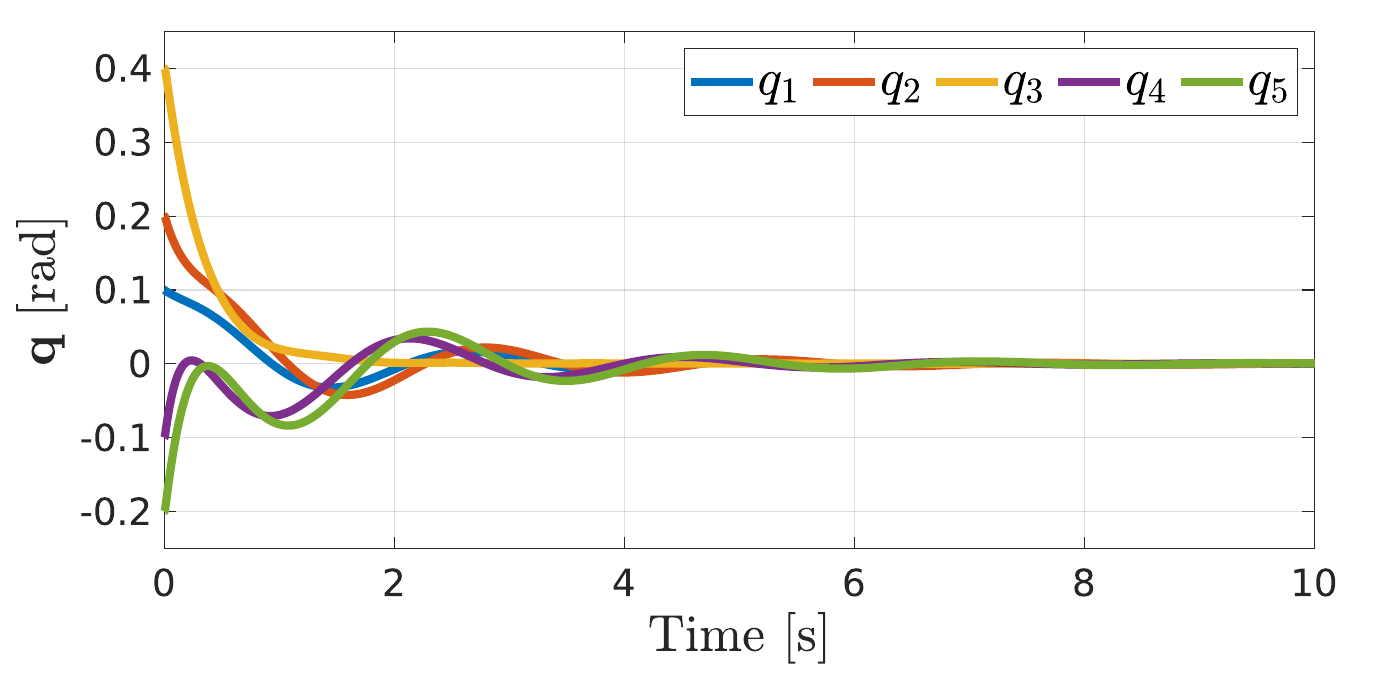}  
  \caption{System behaviour (case A)}
  \label{fig:sub-nom-1}
\vspace{4pt}
\end{subfigure}
\begin{subfigure}{.5\textwidth}
  \centering
\raisebox{4pt}{\includegraphics[width=0.975\linewidth]{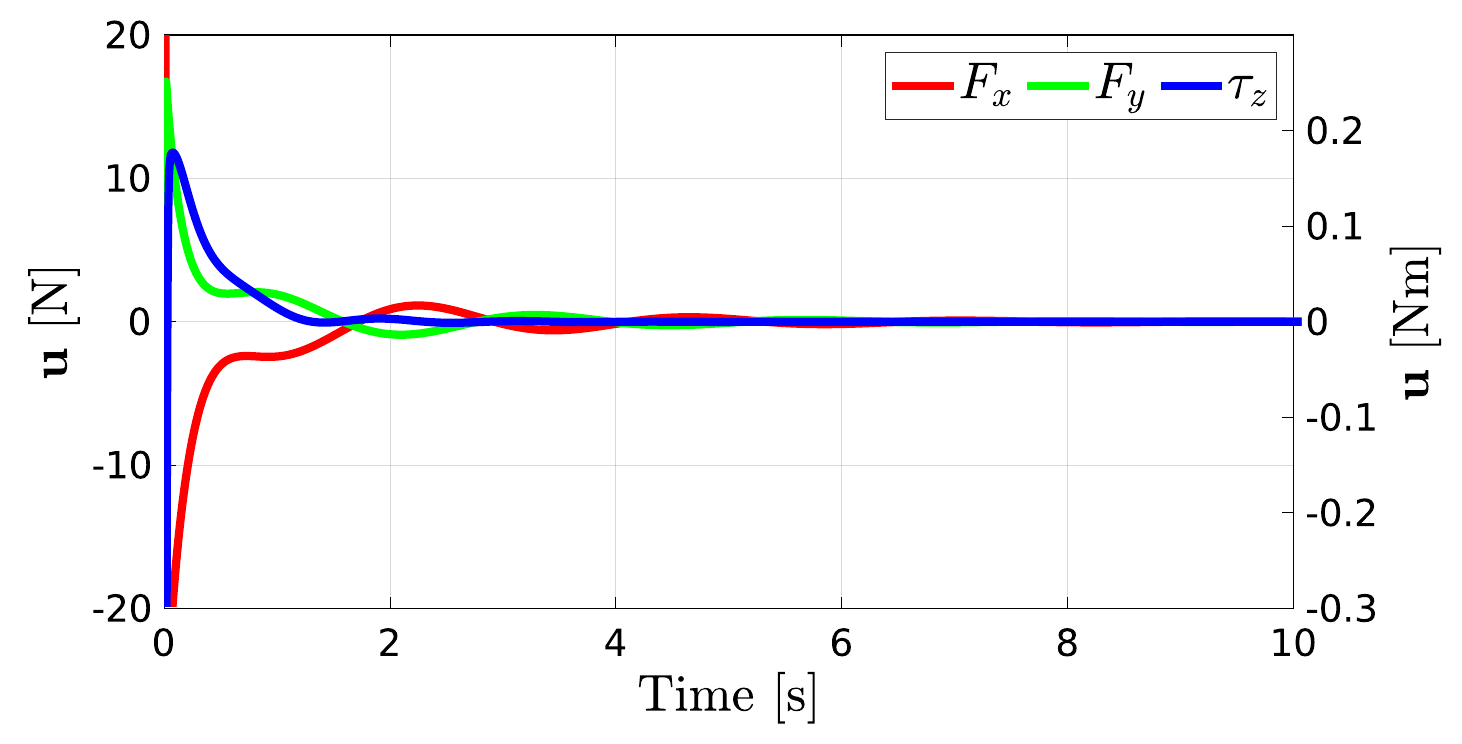}}  
  \caption{Commanded wrenches (case A)}
  \label{fig:sub-nom-2}
\end{subfigure}
\begin{subfigure}{.5\textwidth}
  \centering
  \includegraphics[width=1\linewidth]{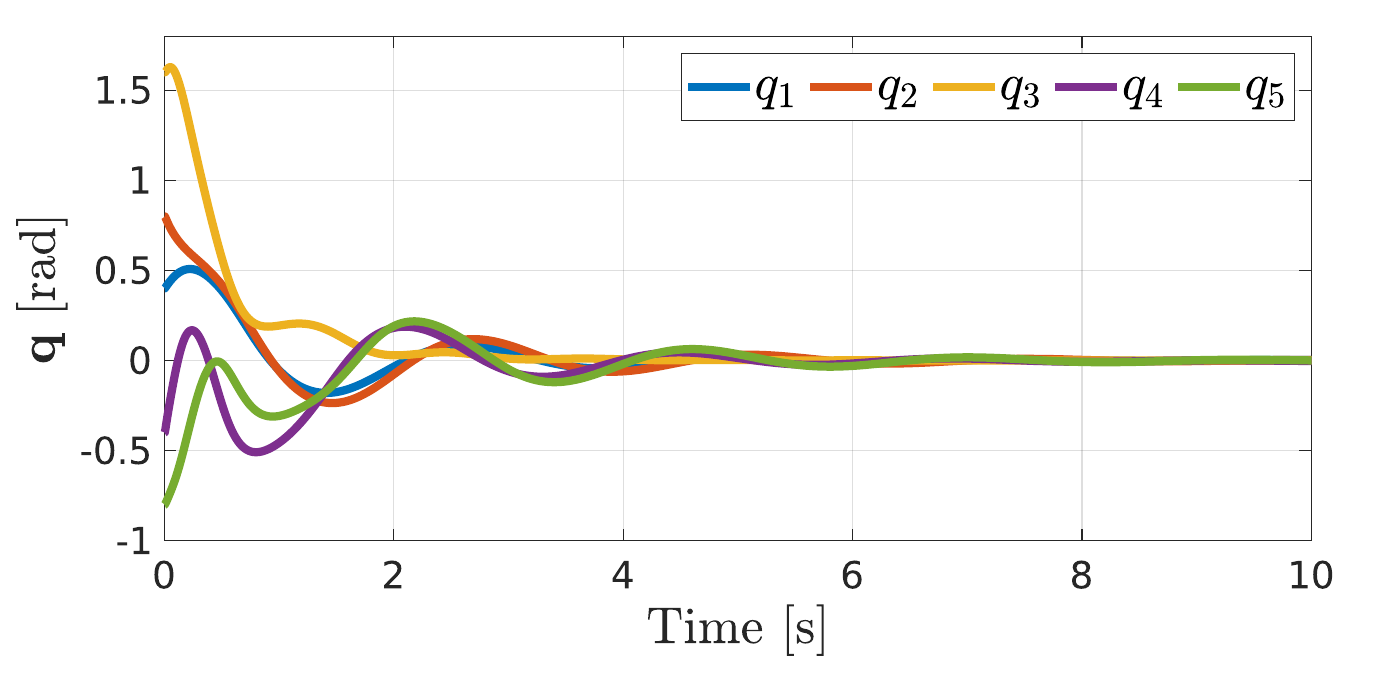}  
  \caption{System behavior (case B)}
  \label{fig:sub-nom-3}
\vspace{4pt}
\end{subfigure}
\begin{subfigure}{.5\textwidth}
  \centering
\raisebox{4pt}{\includegraphics[width=0.975\linewidth]{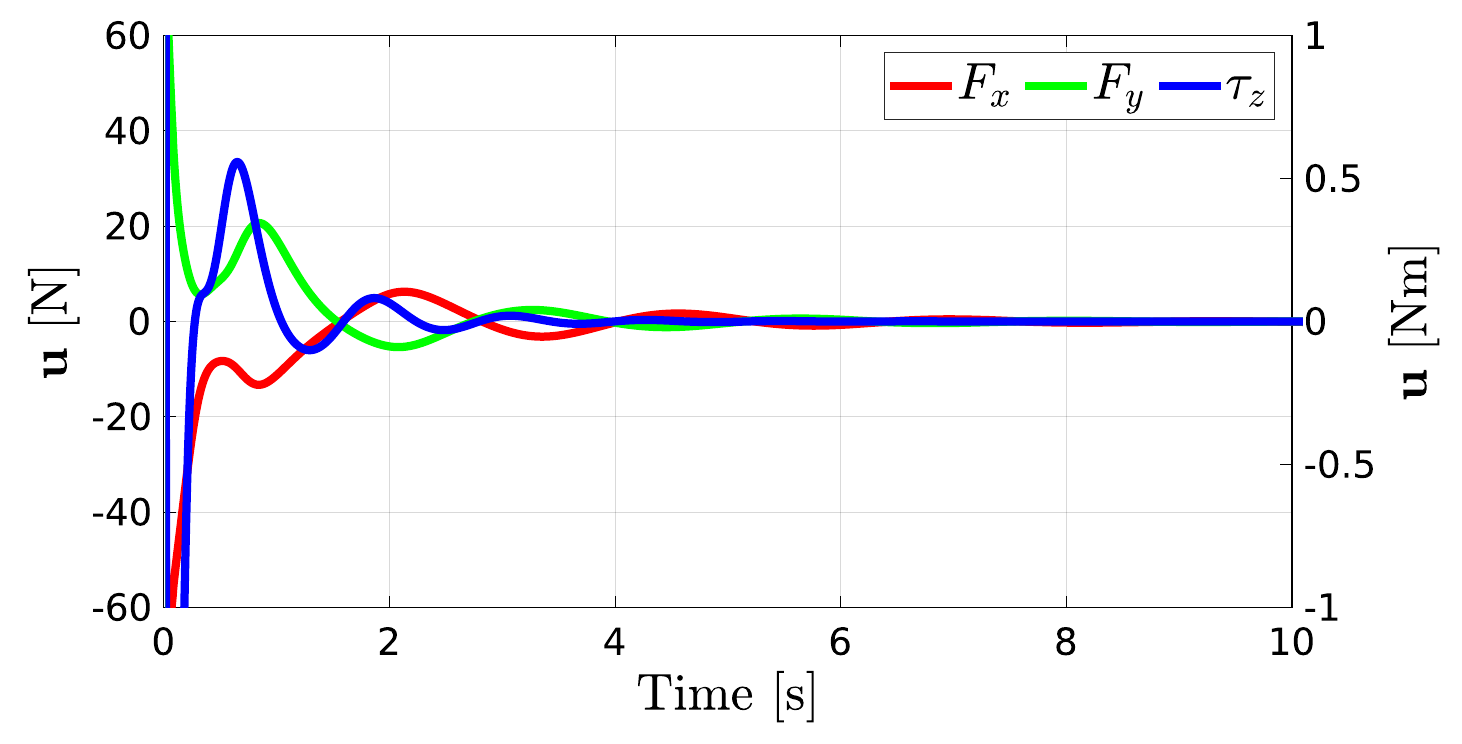}} 
  \caption{Commanded wrenches (case B)}
  \label{fig:sub-nom-4}
\end{subfigure}
\caption{Stabilization of the complete system about its equilibrium for the two nominal cases.}
\label{fig:sim_pend_load}
    \vspace{-15pt}
\end{figure*}

\subsection{Stability Analysis}
In this section, we apply our proposed PFL approach to the 2-DOF planar case of our complete system, and then provide its numerical stability analysis, while comparing it with the standard PFL approach. This simplification allows us to highlight the importance of the coupled terms, with the full model exhibiting similar stability characteristics. Fig. \ref{fig:2DOF_system_beh} shows the stabilization results around
equilibrium from given initial perturbations, using both the approaches, with $\hat{q}_1$ and $\hat{q}_2$ corresponding to the two joints of the planar double pendulum. The joint $\hat{q}_1$ is actuated and considered as the output variable similar to $\mathbf{y}$ in \eqref{eq:dyn_y_pd+}, while the joint $\hat{q}_2$ is passive and constitutes the internal dynamics similar to $\mathbf{q}_c$ in \eqref{eq:dyn_qc}. We consider the parameters of the system dynamics and controller to be the same as those of their corresponding component in the full model, as detailed in the subsequent section. The proposed PFL approach utilizes the control signal for the output dynamics similar to  \eqref{eq:qa_ref_bar}, whereas the standard PFL approach uses the reference similar to \eqref{eq:qa_ref}. The control wrench exerted using the actuated joint $\hat{q}_1$ is calculated similarly to \eqref{eq:u_pd+}. \par 
We observe that for the standard PFL approach, the internal dynamics of $\hat{q}_2$ converge to zero, but the output dynamics of $\hat{q}_1$ enter a stable limit cycle. For our proposed PFL approach, the states corresponding to the internal dynamics and the output dynamics both converge to their equilibrium of zero. Fig. \ref{fig:pfl_com-2} shows the eigenvalues of both these approaches calculated near the equilibrium, and we observe that two of the eigenvalues lie on the imaginary axis for the standard approach, which corresponds to the dynamics of $\hat{q}_1$. By introducing the necessary coupling terms in our proposed law, we observe that these eigenvalues shift towards the left half of the imaginary axis, thereby stabilizing the complete system. Additionally, we also observe that the introduction of the coupling terms has shifted the remaining two eigenvalues corresponding to the dynamics of $\hat{q}_2$, in opposite directions, with both being in the left half of the imaginary axis.

%% file: sections/8-simulations.tex
\label{sec:sim}

In this section, we present the numerical simulation results, first for the nominal case and subsequently for scenarios incorporating different uncertainties. In each case, we demonstrate the stabilization of the complete system \eqref{eq:dyn} about its equilibrium $\left(q_1, q_2, q_3, q_4, q_5\right) =0$.

\subsection{Simulation Setup}
We consider the mass $m_p$ of the underactuated multirotor platform to be \SI{4.06}{\kilogram}, while the mass $m_l$ of the attached load for the nominal case is \SI{1.4}{\kilogram}. The principal moment of inertia of the platform along the $x$, $y$ ad $z$ axis are represented as $I_{xx}$, $I_{yy}$ and $I_{zz}$, respectively, which are considered to be 0.0646, 0.0646, and 0.0682 $\SI{}{\kilogram\square\meter}$, respectively.   The platform is attached from a fixed point using a rigid cable $L_1$ of length \SI{1.5}{\meter}, with the length of cable $L_2$ to attach the load being \SI{0.75}{\meter}. The mass of the cables $L_1$ and $L_2$ are considered as \SI{0.15}{\kilogram} and \SI{0.1}{\kilogram}, respectively. All the numerical simulations are carried out in Matlab/Simulink at a sampling frequency of \SI{1000}{\hertz}.    
 \begin{figure}[t]
    \centering
\includegraphics[width = 0.25\textwidth]{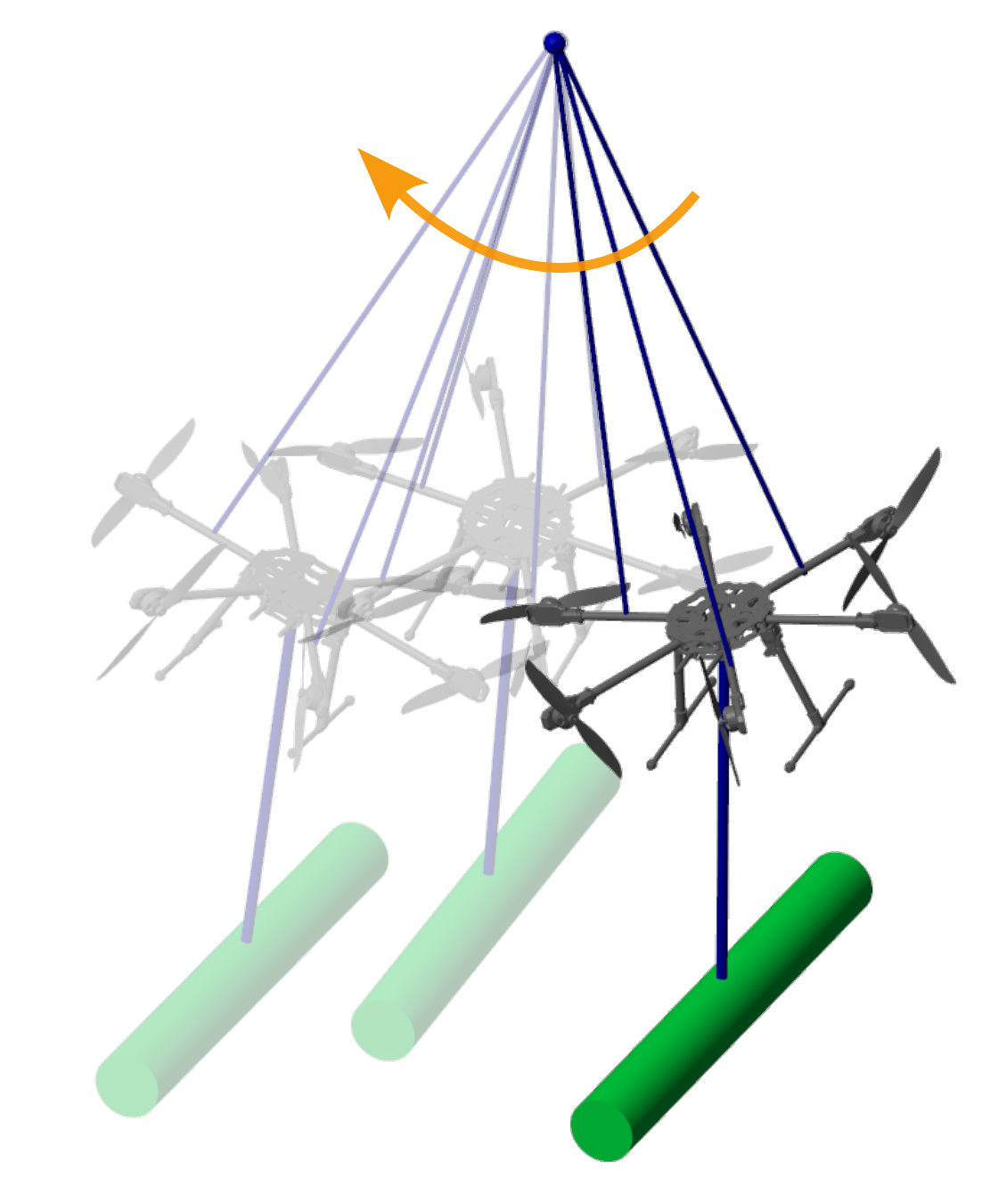}
    \caption{Sequences of tracking a sinusoidal reference by the suspended multirotor platform with the attached load. The orange arrow indicates the direction of the tracking motion.}
    \label{fig:track_sequence}
    \vspace{-20pt}
\end{figure}

\begin{figure*}[t]
\begin{subfigure}{.5\textwidth}
  \centering
\raisebox{0pt}{\includegraphics[width=.95\linewidth]{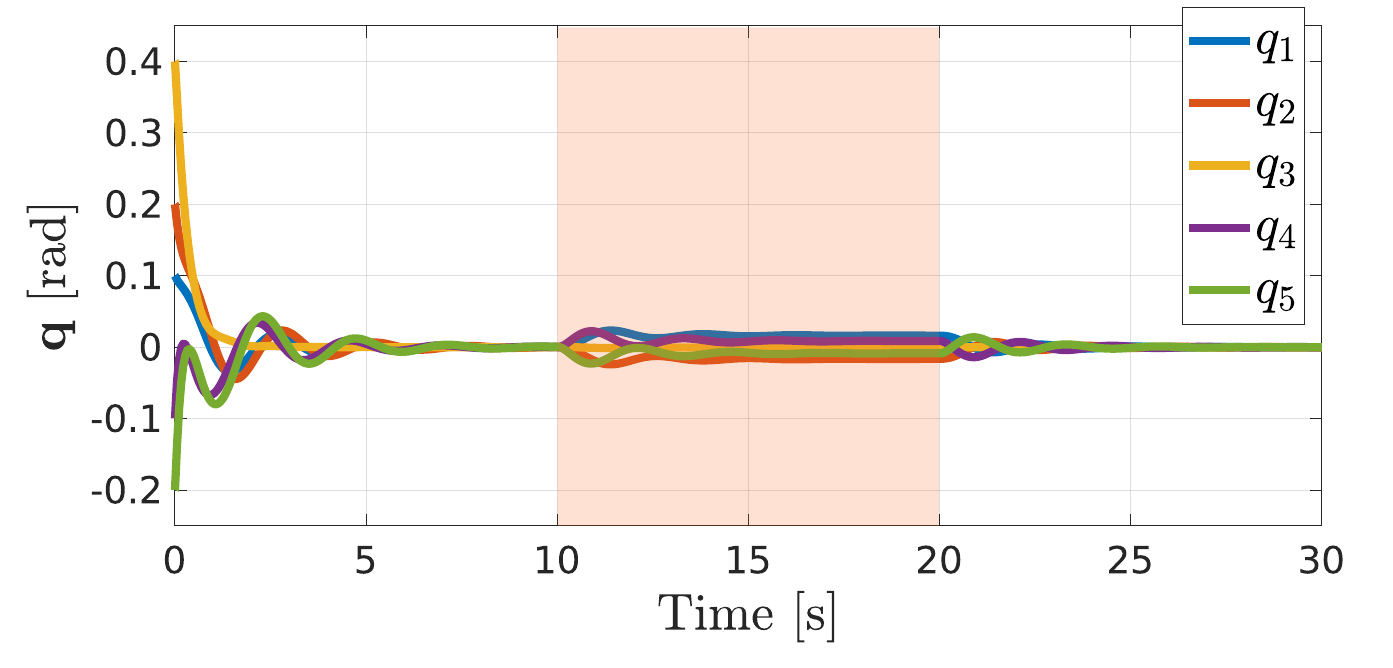}  }
  \caption{System behavior}
  \label{fig:sub-wind-1}
\end{subfigure}
\begin{subfigure}{.5\textwidth}
  \centering
{\includegraphics[width=0.935\linewidth]{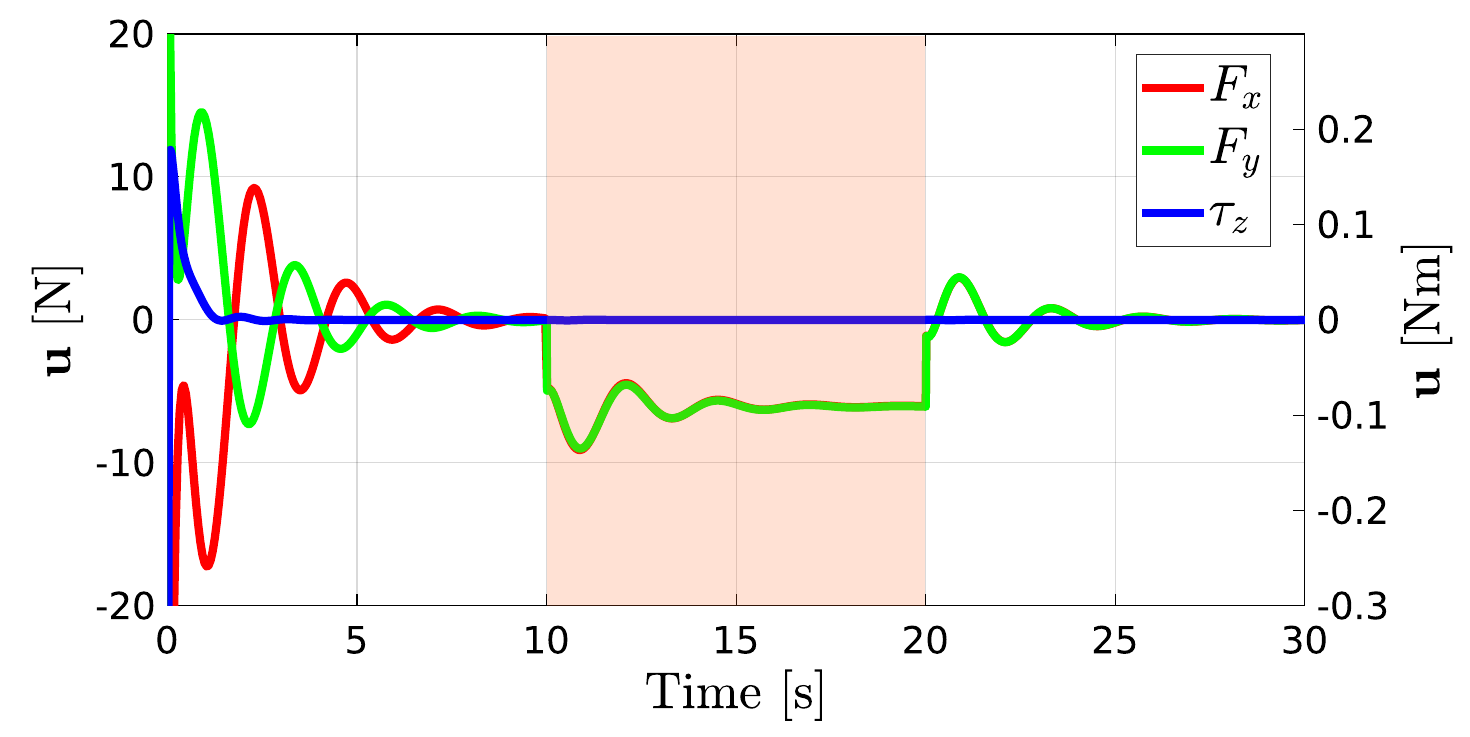} } 
  \caption{Commanded wrenches}
  \label{fig:sub-wind-2}
\end{subfigure}
\caption{System behavior and commanded wrenches by the platform for a heavy load under wind disturbances. The periods of wind disturbance between \SI{10}{\second} and \SI{20}{\second} is highlighted in orange.}
\label{fig:sim_pend_load_wind_load}
\vspace{-10pt}
\end{figure*}

\subsection{Stabilization for Nominal Case}
For the nominal cases, we perturb the system and demonstrate the stabilization results from its initial perturbation, chosen as 0.1, 0.2, 0.4, -0.1, and -0.2 radians for the joints $q_1$, $q_2$, $q_3$, $q_4$, and $q_5$, respectively, for Case A. The corresponding perturbation angles for Case B are chosen as 0.4, 0.8, 1.6, -0.4, and -0.8 radians. The gains of the controller for the output dynamics are chosen as $\mathbf{K}_{py} = diag(4230, 4230, 30)$ and $\mathbf{K}_{dy} = diag(3950, 3950, 10)$, while the control gains for the internal dynamics are chosen as  $\mathbf{K}_{pc} = diag(2200,2200)$ and $\mathbf{K}_{dc} = diag(50,50)$. We observe from Figs. \ref{fig:sub-nom-1} and \ref{fig:sub-nom-3} that our proposed controller has successfully damped the oscillations of the complete system from its given initial states for both cases, from initial angles as high as 86 degrees. Figs. \ref{fig:sub-nom-2} and \ref{fig:sub-nom-4} illustrate the commanded wrenches, showing that the nominal case B demands higher wrench inputs compared to case A, due to its higher initial perturbations. Additionally, we demonstrate that our proposed controller can stabilize the complete system about non-equilibrium points, as shown in Fig. \ref{fig:platform_schematics}.  The provided supplementary video further illustrates this case, while also showcasing the tracking case for a changing reference signal, as depicted in Fig. \ref{fig:track_sequence}. These aspects will be explored in more detail in future work, as they are beyond the scope of this paper, which focuses exclusively on stabilization around equilibrium points.
\par 

\subsection{Robustness Analysis}

In this section, we first analyze the robustness of the proposed control approach to unknown wind disturbances, as it is a key factor affecting the stability and precision of load transportation in our envisioned outdoor crane operations. We considered the external wind disturbance of approximately 10 knots, which impacts the complete system along both the $x$ and $y$ axes. To further enhance the realism of the simulation and better reflect real-world scenarios, we also consider the mass of the attached load to be 20 times the nominal case. While the nominal case provides a standard reference, the increased mass in this case enables us to assess the system’s ability to handle more demanding operating conditions. We observe from Fig. \ref{fig:sub-wind-1} that our proposed PFL controller successfully 
stabilizes the system under the influence of wind disturbances that are exerted between \SI{10}{\second} and \SI{20}{\second}. However, a steady state error of 0.016, -0.016, 0.009, and -0.009 \SI{}{\radian} is observed for the joints $q_1$, $q_2$, $q_4$, and $q_5$, respectively, which can be eliminated by utilizing an external wrench observer \cite{ryll20176d} as future work. The wrenches exerted by the system for the stabilization are shown in Fig. \ref{fig:sub-wind-2}, which demonstrates that the forces $F_x$ and $F_y$ are exerted along the negative $x$ and $y$ axis, respectively,  to counteract the wind disturbances exerted along the positive $x$ and $y$ axis.\par

\begin{table}[h]
    \centering
    \hspace{0cm}\begin{tabular}{|p{1cm}|p{0.5cm}|p{0.5cm}|p{0.75cm}|p{0.75cm}|p{0.75cm}|}
\cline{1-6}
    \multicolumn{1}{|c|}{Cases} &  \multicolumn{1}{c|}{ $m_p$}&  \multicolumn{1}{c|}{$m_l$} &  \multicolumn{1}{c|}{$I_{xx}$   }&  \multicolumn{1}{c|}{$I_{yy}$ } &  \multicolumn{1}{c|}{$I_{zz}$  }\\
\Xhline{3\arrayrulewidth}  
        Nominal  & 4.06 & 1.4 & 0.0646 & 0.0646 & 0.0682 \\
       \hline 
          Uncertain  & 10.06 & 20.4 & 0.75 & 0.75 & 0.5 \\
       \hline 
    \end{tabular}
    \caption{Model mismatch parameters for the robustness analysis. The mass is measured in \SI{}{\kilogram}, whereas the moment of inertia is measured in \SI{}{\kilogram\square\meter}.}
    \label{tab:model_mismatch}
\end{table}
\begin{figure}[t]
\begin{subfigure}{.5\textwidth}
  \centering
\includegraphics[height=0.475\linewidth]{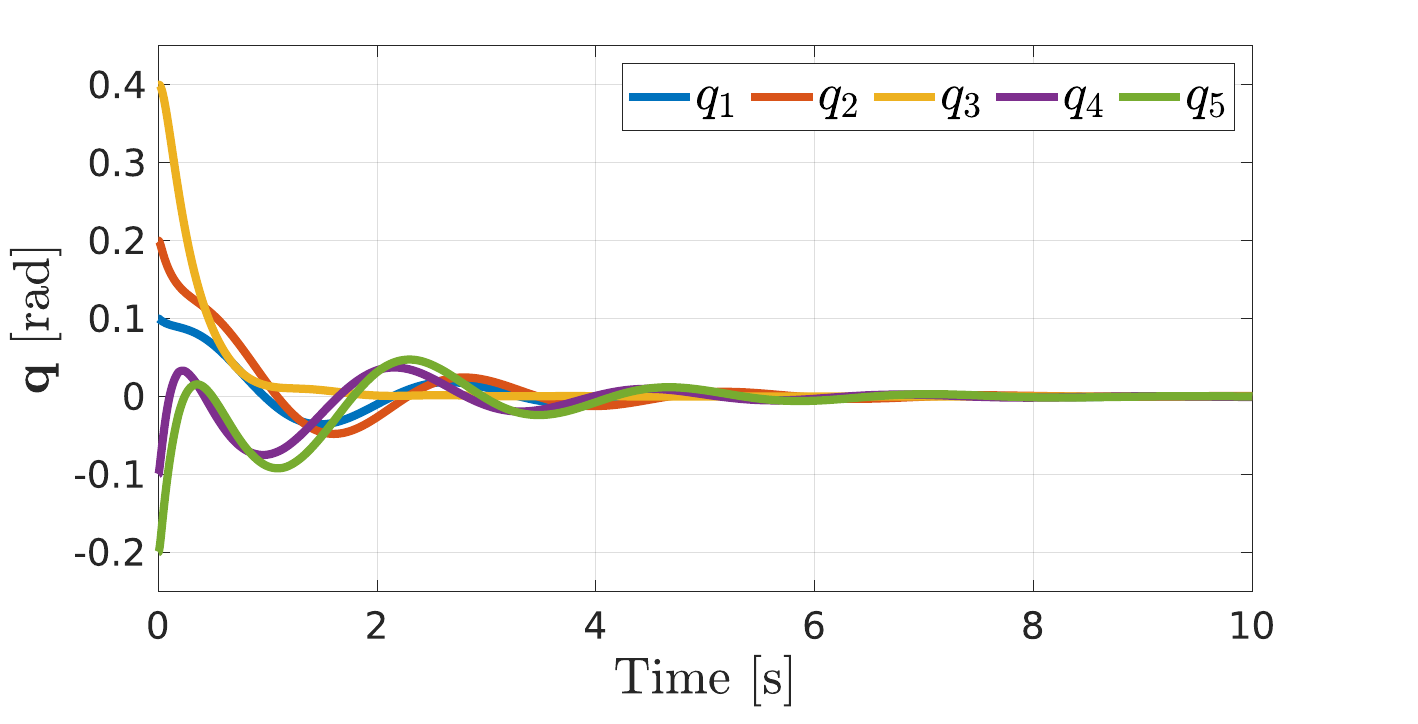}  
  \caption{System behaviour}
  \label{fig:sub-robust-1}
\end{subfigure}
\begin{subfigure}{.5\textwidth}
  \centering
\includegraphics[width=.95\linewidth]{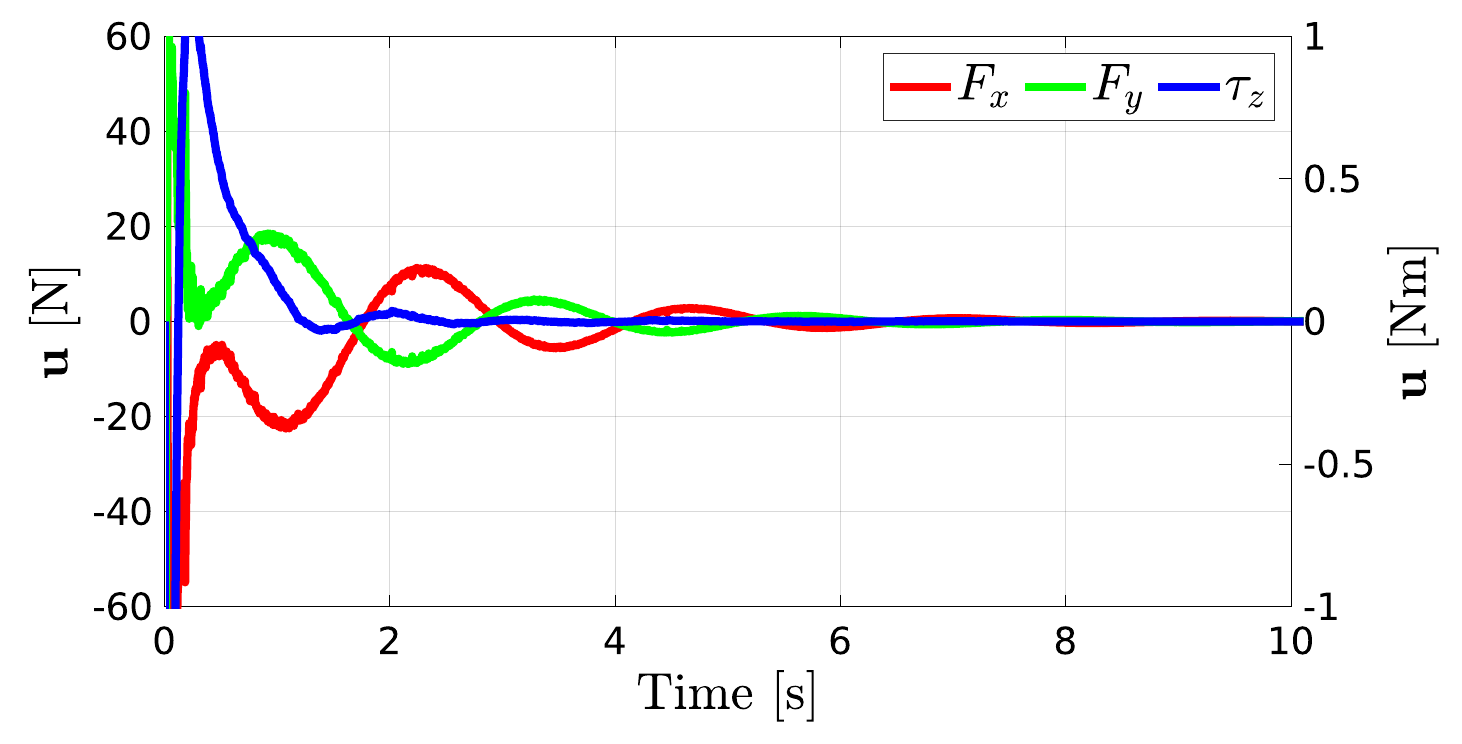}  
  \caption{Commanded wrenches}
  \label{fig:sub-robust-2}
\end{subfigure}
\caption{System behaviour and commanded wrenches by the platform for the stabilization scenario consisting of model uncertainties and sensor noise.}
\label{fig:sim_pend_load_robust}
\vspace{0pt}
\end{figure}
   \begin{table}[t]  
  \centering
\hspace{2cm} \begin{tabular}{|p{1cm}|p{0.5cm}|p{0.4cm}|p{0.5cm}|p{0.85cm}|p{0.5cm}|p{0.5cm}|p{0.6cm}|}
    \cline{2-8}
    \multicolumn{1}{c|}{} &  \multicolumn{2}{c|}{Response }&  \multicolumn{2}{c|}{Peak } &  \multicolumn{3}{c|}{SNR  }\\
     \multicolumn{1}{c|}{} &  \multicolumn{2}{c|}{Time (s)}&  \multicolumn{2}{c|}{Response (rad)} &  \multicolumn{3}{c|}{(db)} \\     \cline{1-8}
    \multicolumn{1}{|c|}{Cases} &  $q_4$ & $q_5$ & $q_4$ & $q_5$ & $F_x$ & $F_y$ & $\tau_x$\\ 
    \Xhline{3\arrayrulewidth}  
      Nominal & 8.33 & 8.6 & 0.005 & -0.003 & 242.1 & 238.7 & 210.9\\\hline 
     Uncertain & 8.27 & 8.6 &  0.036 & 0.018 & 29.6 & 26.9 & -4.8\\\hline 
  \end{tabular}
  \caption{KPIs to analyze the robustness of our proposed approach.}
    \label{tab:kpi_robust_nominal}
    \vspace{-10pt}
\end{table}

Next, we analyze the robustness of our proposed approach to different 
uncertainties, such as variation in the mass and moment of inertia of the suspended platform, as summarized in Table \ref{tab:model_mismatch}. We also vary the mass of the attached load for the robustness analysis to more than $90\%$ of its nominal mass. Additionally, we introduce noise to the acceleration measurement. The simulated noise is zero-mean Gaussian noise with a variance of \SI{1}{\meter\second^2}, with the strength of the noise being $10 \%$ of the actual signal.  It is important to note that these uncertainties are not accounted for in the design of the controller.  As observed from Fig. \ref{fig:sim_pend_load_robust}, our control approach has successfully stabilized the complete system in the presence of these involved uncertainties. 
 \par

 \begin{figure*}[t]
    \centering
\includegraphics[width = 0.95 \textwidth]{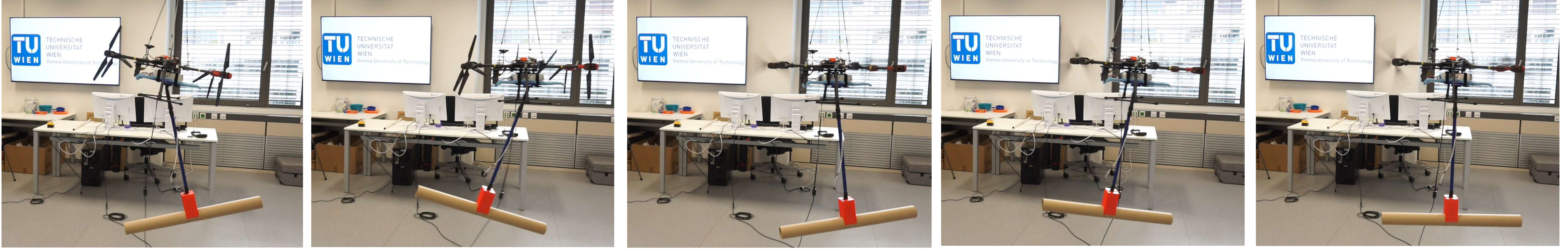}
    \caption{{Sequences of the experiments for stabilizing the suspended aerial platform with the attached known load.}}
    \label{fig:exp_sequence_2}
\end{figure*} 
 \begin{figure*}[h]
    \centering
\includegraphics[width = 0.95 \textwidth]{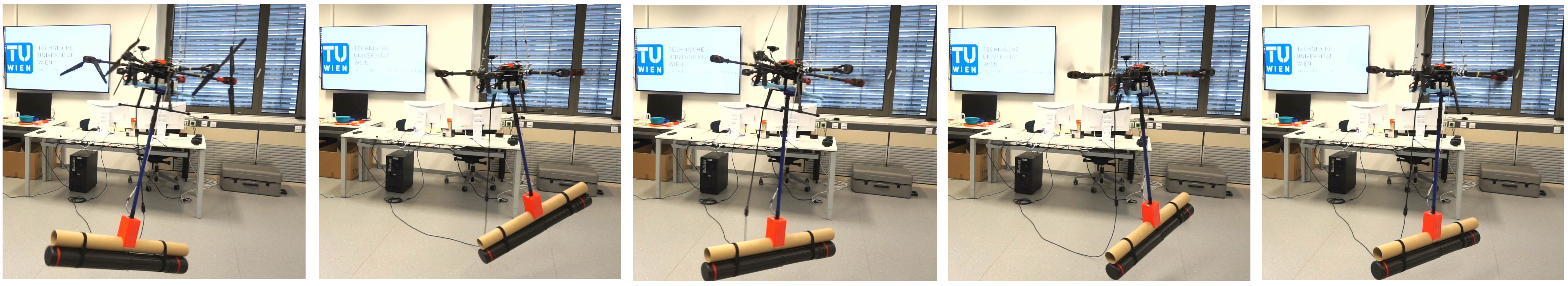}
    \caption{Sequences of the experiments for stabilizing the suspended aerial platform with the attached unknown load. }
    \label{fig:exp_sequence_2}
    \vspace{-10pt}
\end{figure*} 

We have introduced the three key performance indicators (KPIs), summarized in the Table. \ref{tab:kpi_robust_nominal}, in order to evaluate the robustness of our controller and compare it with the nominal case. The peak response represents the largest overshoot observed in the first transient phase before equilibrium is approached, while response time is the time required for the response to reach $1\%$ of its steady-state value. Finally, the signal-to-noise ratio (SNR) is a measure of the noise in the control input \cite{box1988signal}, where a higher SNR denotes a lower noise level, and vice versa. We have analyzed the KPIs of the load states $q_4$ and $q_5$, as the KPIs for the other states show comparable results for both the nominal and uncertain case. 
We observe that due to the influence of the sensor noise, the SNR of the commanded wrench in this case is around 10 times lower than in the nominal case. However, the presence of sensor noise and model uncertainties did not impact the response time of the load dynamics. When compared with the nominal case, a slight increase in the peak response is observed for the joints $q_4$ and $q_5$ by 0.031 m and 0.015 m, respectively.

%% file: sections/9-experiments.tex
\label{sec:expm}

Our proposed control approach is validated in real experiments using the planar-thrust platform fitted with 15-inch propellers, used in our previous work \cite{das2023, das2024}. 
We generate the desired wrench using our proposed controller \eqref{u:proposed}, and send it to the platform by issuing the corresponding PWM signals to the brushless DC motors. The states of the platform and the attached load are obtained using our designed state estimation modules \cite{das2023, das2024}, which use the onboard Pixhawk 2.1 flight controller unit (FCU) on the platform, and the MPU-9250 inertial measurement unit (IMU) attached to the load for its observation vector.  Although the state estimator accounts for sensor noise through its associated covariances, residual uncertainties due to parameter variations and electronic interference remain, against which our designed controller will be tested for robustness. \par 
The goal of the experiment is to stabilize the complete system about its equilibrium, from a given initial configuration in two different scenarios, first with a known attached load and then with an unknown attached load to test the robustness of our approach. 
\begin{figure}[h]
\begin{subfigure}{.5\textwidth}
  \centering
  \includegraphics[width=.95\linewidth]{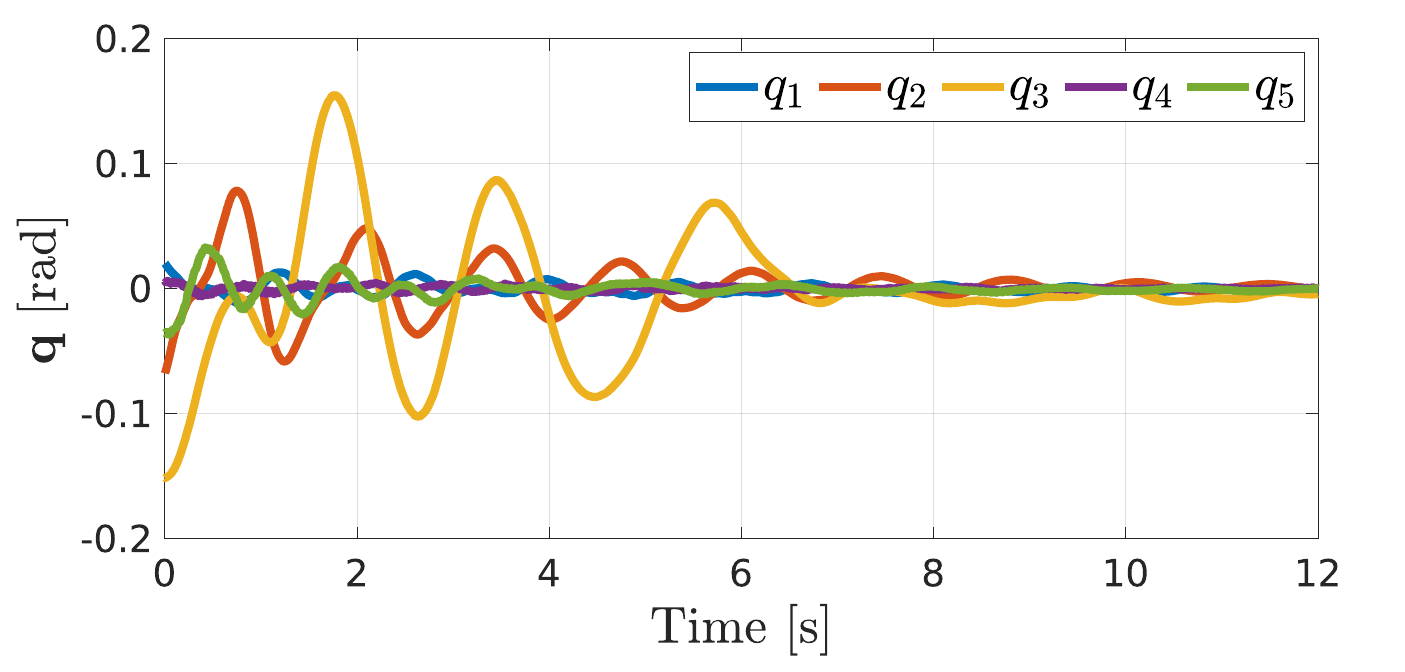}
  \caption{}
  \label{fig:exp_pend_load_q}
\end{subfigure}
\begin{subfigure}{.5\textwidth}
  \centering
\includegraphics[width=0.95\linewidth]
{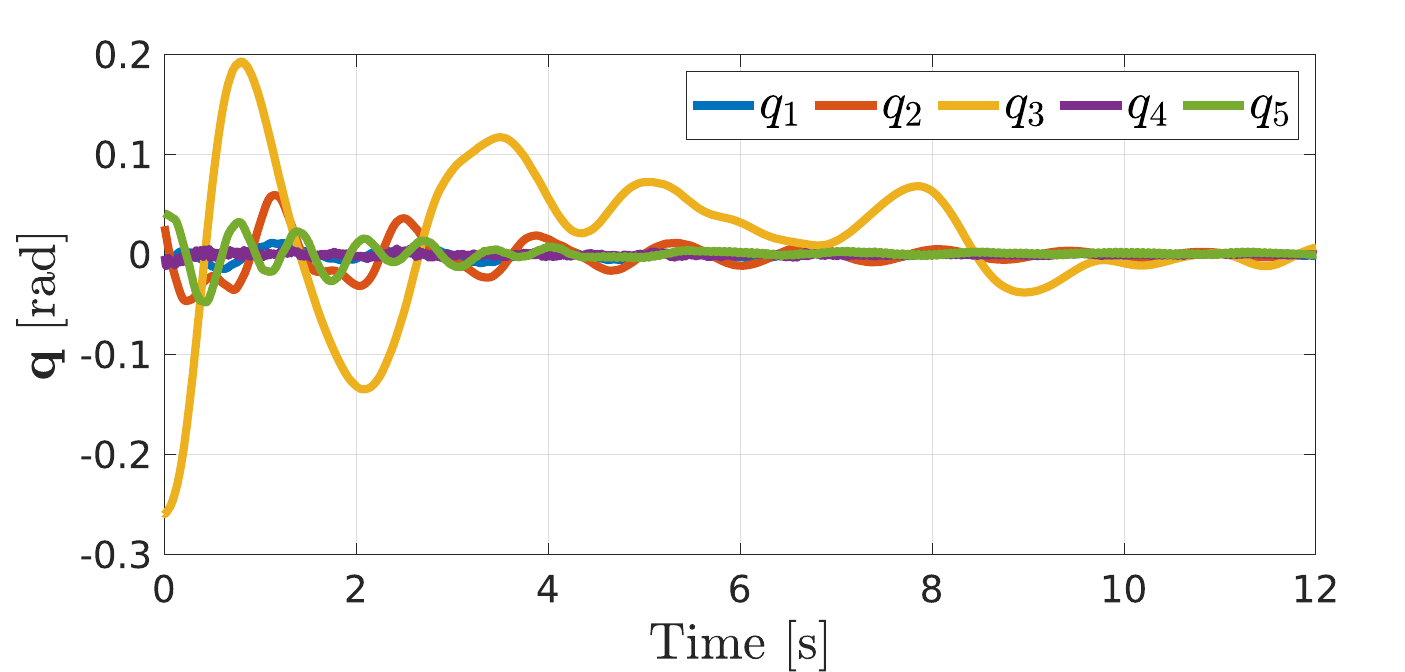}  
\caption{}
    \label{fig:exp_pend_load_q_robust}
\end{subfigure}
\caption{{Experimental results showing the system behavior of the suspended platform with the attached (a) known load and (b) unknown load.}}
\label{fig:exp}
\vspace{-10pt}
\end{figure}
The experimental result for the nominal case with a known attached load is shown in Fig. \ref{fig:exp_pend_load_q}. Note that to reduce the model dependency on the proposed PFL controller, the gravity and Coriolis compensation are not considered in both control laws \eqref{eq:u_pd+} and \eqref{eq:qa_ref_bar}, which are not necessary for stabilization of the complete system about its equilibrium. The control gains are chosen based on the nominal moment of inertia and mass of the complete system as  $\mathbf{{K}}_{py}=$ diag$\left(18, 18, 0.35 \right)$, $\mathbf{{K}}_{pc}=$ diag$\left(6, 6 \right)$, $\mathbf{{K}}_{dy}=$ diag$\left(13, 13, 0.15 \right)$ and $\mathbf{{K}}_{dc}=$ diag$\left(0.25, 0.25 \right)$. \par 
We observe that all the states have converged to their equilibrium with a satisfactory time of around \SI{10}{\second}, with a relatively slow convergence rate for the joint $q_3$, as its dynamics is relatively faster than the maximum time-constant of its corresponding applied wrench $\tau_z$, which is not the case for the other states. Note that a faster convergence of the attached load due to the chosen set of gains will further lead to a faster convergence of the suspended platform, and vice-versa.  
The experimental results for the other scenario are shown in Fig. \ref{fig:exp_pend_load_q_robust}, where an unknown load of mass \SI{0.45}{\kilogram} is attached to the suspended platform. We observe that the proposed controller is robust to the uncertainty in the mass of the attached load, with a response time comparable to the nominal case.

\par

%% file: sections/10-conclusion.tex
\label{sec:con}
We presented a novel control approach for the stabilization of a suspended multirotor platform with an attached load. Our approach is based on partial feedback linearization, which considers the underactuation of the complete system by incorporating certain coupled terms into the stabilizing control law of the output dynamics. Through numerical stability analysis, we demonstrated that these coupled terms are crucial in the control law. Furthermore, our proposed control approach was shown
to be robust to unmodelled system dynamics, sensor noise, and external wind disturbances, as validated by both extensive simulation studies and experimental tests. Additionally, our approach relied on only onboard sensors for stabilization, making it well-suited for its intended application in construction sites.  \par 
In the future, we plan to incorporate a nonlinear observer for the estimation of wind disturbances and incorporate it in our control law. Additionally, we aim to combine our proposed approach with an active crane while demonstrating stabilization and transportation of heavy loads.